%% file: paper.tex
\documentclass[10pt,twocolumn,letterpaper]{article}

\usepackage[accsupp]{axessibility}
\usepackage{cvpr}              

\usepackage{graphicx}
\usepackage{amsmath}
\usepackage{amssymb}
\usepackage{booktabs}
\usepackage{multirow}
\usepackage[linesnumbered,ruled,vlined]{algorithm2e}

\usepackage[pagebackref,breaklinks,colorlinks]{hyperref}

\usepackage[capitalize]{cleveref}
\crefname{section}{Sec.}{Secs.}
\Crefname{section}{Section}{Sections}
\Crefname{table}{Table}{Tables}
\crefname{table}{Tab.}{Tabs.}

\def\cvprPaperID{3953} 
\def\confName{CVPR}
\def\confYear{2022}

    \makeatletter
\def\@fnsymbol#1{\ensuremath{\ifcase#1\or \ddagger\or \ddagger\or
   \mathsection\or \mathparagraph\or \|\or **\or \dagger\dagger
   \or \ddagger\ddagger \else\@ctrerr\fi}}
    \makeatother
    
\begin{document}

\title{RSCFed: Random Sampling Consensus Federated Semi-supervised Learning}
\author{Xiaoxiao Liang$^1$, Yiqun Lin$^1$, Huazhu Fu$^2$, Lei Zhu$^{3, 1}$, Xiaomeng Li$^1$\thanks{Project lead and corresponding author.} \\
$^1$~The Hong Kong University of Science and Technology, 
$^2$~IHPC, A*STAR \\
$^3$~The Hong Kong University of Science and Technology (Guangzhou)\\ 
{\tt\small \{xliangak, ylindw\}@connect.ust.hk, hzfu@ieee.org, \{leizhu, eexmli\}@ust.hk}}
\maketitle

\newcommand{\xmli}[1]{{\color[rgb]{0,0,0}{#1}}}
\newcommand{\hz}[1]{{\color[rgb]{0,0,1}{[hz:#1]}}}

\newcommand{\yq}[1]{{\color[rgb]{0.1,0.1,0.1}{[#1]}}}
\newcommand{\yqs}[1]{{\color[rgb]{0.7,0.2,0.1}{[\sout{#1}]}}}
\newcommand{\xx}[1]{{\color[rgb]{1,0,1}{#1}}}

\input{Section0-abstract}
\input{Section1-introduction}
\input{Section2-relatedwork}
\input{Section3-method}
\input{Section4-Experiments}
\input{Section5-conclusion}

\section*{Acknowledgement}
This work was supported by  a research grant from HKUST Bridge Gap Fund (BGF.027.2021), a research grant from Shenzhen Municipal Central Government Guides Local Science and Technology Development Special Funded Projects (2021Szvup139), a research grant from the National
Natural Science Foundation of China (Grant No. 61902275), and A*STAR AI3 HTPO Seed Fund (C211118012).


\bibliographystyle{ieee_fullname}
\bibliography{ref}

\end{document}

%% file: Section0-abstract.tex
\begin{abstract}


\xmli{Federated semi-supervised learning (FSSL) aims to derive a global model by training fully-labeled and fully-unlabeled clients or training partially labeled clients.} The existing approaches work well when local clients have independent and identically distributed (IID) data but fail to generalize to a more practical FSSL setting, \textit{i.e.}, Non-IID setting.
In this paper, we present a \textbf{R}andom \textbf{S}ampling
\textbf{C}onsensus \textbf{Fed}erated learning, namely \textbf{RSCFed}, by considering the uneven reliability among models \xmli{from fully-labeled clients, fully-unlabeled clients or partially labeled clients.}
Our key motivation is that given models with large deviations from either labeled clients or unlabeled clients, the consensus could be reached by performing random sub-sampling over clients. 
To achieve it, instead of directly aggregating local models, we first distill several sub-consensus models by random sub-sampling over clients and then aggregating the sub-consensus models to the global model. 
To enhance the robustness of sub-consensus models, we also develop a novel distance-reweighted model aggregation method. 
Experimental results show that our method outperforms state-of-the-art methods on three benchmarked datasets, including both natural and medical images. The code is available at \href{https://github.com/XMed-Lab/RSCFed}{https://github.com/XMed-Lab/RSCFed}.

\end{abstract}

%% file: Section1-introduction.tex
\section{Introduction}

The core idea of federated learning (FL) is to train machine learning models on separate datasets that are distributed across different places or devices, which can preserve local data privacy to a certain extent. Over the past few years, FL has emerged as an important research area and attracted many researchers' attention to study its application in medical image diagnosis~\cite{kaissis2020secure,yang2021federated,kumar2021blockchain}, image classification~\cite{li2021model} and object detection~\cite{liu2020fedvision}.

Considerable efforts have been devoted to design various FL methods, such as FedAvg~\cite{mcmahan2017communication}, SCAFFOLD~\cite{karimireddy2020scaffold} and MOON~\cite{li2021model}. Although the results are quite promising, these methods require fully labeled images on each local client, limiting its application in real practice. 



\begin{figure}[!t]
    \includegraphics[width=\linewidth]{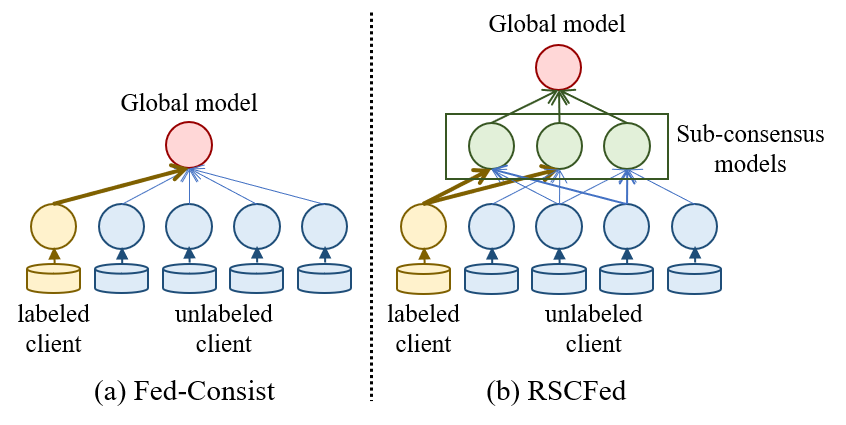}
    \vspace{-7mm}
    \caption{Illustration of the existing FSSL method, \textit{i.e.},~\cite{yang2021federated} and our RSCFed. Existing methods simply perform the standard model aggregation in FedAvg~\cite{mcmahan2017communication}, while our RSCFed distills multiple sub-consensus models from local models and updates the global model via aggregating sub-consensus models.} 
    \label{intro}
    \vskip -15pt
\end{figure}

Recently, federated semi-supervised learning (FSSL)~\cite{yang2021federated,liu2021federated,jeong2021federated,lin2021semifed} is becoming a new research topic, aiming at utilizing the unlabeled images to enhance the global model development. One line of the research studies FSSL by considering each client has partially labeled and unlabeled images. For example, Jeong~\etal~\cite{jeong2021federated} introduced inter-client consistency loss to improve the global model by encouraging the consistent outputs from multiple clients. Another line of FSSL~\cite{yang2021federated,liu2021federated} assumes that some local clients have fully labeled images while some clients contain unlabeled images, which we denote as labeled clients and unlabeled clients respectively.
However, existing methods have two main limitations. 
First, they do not consider not independent and identically distributed data (Non-IID) among local clients, which is a key problem for FL and can cause a deterioration in accuracy~\cite{kairouz2019advances,li2021federated}.
Second, some solutions~\cite{liu2021federated} share the correlation matrix among local clients,
which might cause information leakage. 

\if 
To solve this problem, Yang~\etal~\cite{yang2021federated} applied standard supervised and unsupervised training methods to labeled and unlabeled clients, respectively and obtain the final global model via FedAvg~\cite{mcmahan2017communication}. 
Liu~\etal~\cite{liu2021federated} further improves FSSL by incorporating the inter-class relationship as a way to switch the information between labeled and unlabeled clients. 
\fi


\begin{figure}[!t]
    \includegraphics[width=\linewidth]{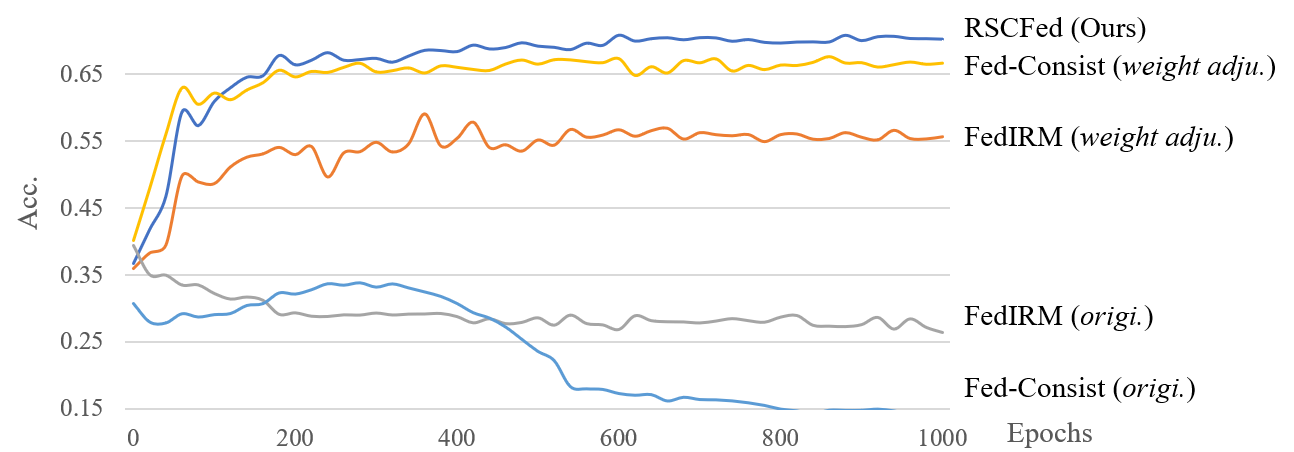}
    \vspace{-7mm}
    \caption{Comparisons of test accuracy curve of our RSCFed with FedIRM~\cite{liu2021federated} and  Fed-Consist~\cite{yang2021federated} under original setting (\emph{origi.}) and under our weight adjusting setting (\emph{weight adju.}).}
    \label{SVHN_compare}
    \vskip -15pt
\end{figure}
\par





\xmli{This paper studies the FSSL with two widely used settings: (1) jointly training fully-labeled and fully-unlabeled clients; (2) jointly training partially-labeled clients.}
A straightforward solution is to extend existing FSSL methods~\cite{liu2021federated, yang2021federated} to the Non-IID setting. However, both FedIRM~\cite{liu2021federated} and Fed-Consist~\cite{yang2021federated} fail to generalize to the Non-IID setting.  
This is because FedIRM~\cite{liu2021federated} proposed to share an inter-class correlation matrix among clients \xmli{based on the assumption that each client has the same class relationship}. \xmli{However, the class relationship can not be correctly learned due to the heterogeneous data among local clients in the Non-IID setting, thus hurting the model performance.} Fed-Consist~\cite{yang2021federated} proposed to equally average the model weights from labeled and unlabeled clients. However, the performance significantly decreases when unlabeled clients increases since the global model may be dominated by the unlabeled clients. Adjusting aggregation weights for labeled and unlabeled clients is one solution, \ie, increasing the weights for labeled clients while decreasing the weights for unlabeled ones. Nevertheless, this result achieves limited performance; see \emph{(weight adju.)} in Fig.~\ref{SVHN_compare}.

\par 

To this end, we present \textbf{R}andom \textbf{S}ampling \textbf{C}onsensus \textbf{Fed}erated learning, namely \textbf{RSCFed}, by considering the {\bf \emph{uneven reliability}} \xmli{among models either from fully-labeled and fully-unlabeled clients or from several partially-labeled clients under the Non-IID setting} without {\bf \emph{any information leakage among clients}}.~\xmli{For example}, labeled clients are easily biased to local data, while unlabeled clients are difficult to achieve the high accuracy, leading to uneven model reliability among clients. \xmli{On the other hand, training with several partially-labeled clients may also cause uneven model reliability because images in each client are heterogeneously distributed in quantity skew and label skew.}
To achieve a robust global model, our key idea is \emph{to regard the local models as noisy models and distill several consensus models via random sampling before aggregating to the global model}, as shown in Fig.~\ref{intro}. 
Specifically, in each synchronization round, we randomly sub-sample clients and record the averaged weights from the sub-sampled models as \emph{\textbf{a sub-consensus model.}}
By performing the operation multiple times, we update the global model via aggregating multiple sub-consensus models. 
To distill a robust sub-consensus model from randomly sampled local clients, we introduce a distance-reweighted model aggregation (DMA) module, which dynamically increases the weights for models that are close to the sub-consensus model and vice versa. 
The idea shares a similar spirit with random sample consensus (RANSAC)~\cite{fischler1981random}, which identifies points as outliers if they are far from the model. 
We conduct extensive experiments on natural image classification datasets (e.g., SVHN and CIFAR-100) and medical dataset (i.e., ISIC 2018 Skin) to demonstrate the effectiveness of RSCFed. Overall, our main contributions can be summarized as follows:
 
\if 1 
We also demonstrate the effectiveness of RSCFed and conclude that more unlabeled clients involved, the better results RSCFed can achieve. 

We conduct extensive experiments on natural image classification dataset including SVHN and CIFAR-100 to evaluate the effectiveness of FedRSC. To better address the practicality of our FSSL scheme, we simulate the realistic computer-assisted diagnosis scenario, where many hospitals are unable to annotate clinical images or diagnose accurately due to lack of budget and experts. Therefore, we also evaluate our method on medical dataset, ISIC 2018: Skin Lesion Analysis Towards Melanoma Detection. FedRSC significantly outperforms the other state-of-the-art FSSL methods in most cases by at least 2\% accuracy.
\fi

\begin{itemize}
\setlength{\parskip}{-0pt}
\setlength{\leftmargin}{-15pt}
\vskip -10pt
\item \xmli{In this paper, we present a novel FSSL method, named RSCFed, to address the uneven reliability of Non-IID local clients. Unlike existing  FSSL frameworks that directly aggregate local clients, RSCFed proposes the concept of updating the global model via aggregating multiple sub-consensus models.}
  
    
\item 
   To improve the sub-consensus model, we introduce a novel distance-reweighted model aggregation (DMA) module, which dynamically adjusts the weights of each sampled local client to the sub-consensus model. 
   

    \item  Experiments on three public datasets demonstrate that our RSCFed significantly outperforms the other state-of-the-art FSSL methods. We further show that with larger ratio of unlabeled data  involved, the better improvement RSCFed can achieve. 
\end{itemize}

%% file: Section2-relatedwork.tex
\section{Related Work}
\subsection{Federated Learning with Non-IID}
Federated learning provides multi-institutional data collaboration solutions for model training under a data-decentralized scheme~\cite{konevcny2016federated,yang2019federated}. Two common problems in this field are system heterogeneity and statistical heterogeneity, which refer to the inconsistency of computational abilities and data distribution among clients. 
A pioneering work provided the most widely recognized FL baseline, FedAvg~\cite{mcmahan2017communication}, followed by many heterogeneous FL solutions, which could be categorized into two branches: local training-oriented methods~\cite{li2021model} and model aggregation-oriented methods.

\noindent \textbf{Local Training-oriented Methods} As for local training-oriented methods, Li~\etal~\cite{li2018federated} add an additional regularization term in local objectives, representing the distance between the global model and local model, thus giving constraints on the model drift. Besides, Karimireddy~\etal~\cite{karimireddy2020scaffold} prove control variates to correct local model update, and Li~\etal~\cite{li2021model} introduce a contrastive loss term to prevent local models from their local minimum. Several other methods perform inter-client privacy-invariant information exchange~\cite{yoon2021fedmix,liu2021feddg}. 
However, most existing methods for Non-IID data fail under the FSSL setting due to the uneven model reliability from labeled and unlabeled clients. 
Besides, some methods exchange the information among clients~\cite{yoon2021fedmix,liu2021feddg}, which may have the potential for information leakage. Unlike these methods, we do not share any information among clients.

\noindent \textbf{Model Aggregation-oriented Methods} As for improvements on model aggregation, Wang~\etal~\cite{wang2020tackling} normalize the received local gradients before averaging;  Wang~\etal~\cite{Wang2020Federated} perform layer-wise averaging with Bayesian non-parametric methods; Chen~\etal~\cite{chen2021fedbe} regard the known global and local models as samples from an assumed distribution, where another set of models are sampled as teacher models and are later utilized in server-side knowledge distillation under the assumption that unlabeled data could be kept at the server. Zhang~\etal~\cite{zhang2021personalized} further extends a single global model to multiple global models, in which affinity towards all global model candidates are computed at each client. Finally, global models are weighted averaged by affinity in a personalized manner according to each client. 
However, these methods are developed for supervised federated learning, while our work focuses on FSSL with uneven model reliability from labeled and unlabeled clients.


\subsection{Semi-Supervised Learning}
Standard semi-supervised learning aims to optimize a model with both labeled and unlabeled data in a centralized manner. The learning paradigm usually involves smoothness-based consistency regularization~\cite{tarvainen2017mean,yu2019uncertainty,li2020transformation,cui2019semi}, entropy minimization-based self-training methods, \cite{chen2021semi,zou2020pseudoseg}, and their combinations~\cite{berthelot2019mixmatch,sohn2020fixmatch,hu2021simple}. For instance, Zou~\etal~\cite{zou2020pseudoseg} fuse the decoder prediction and self-attention Grad-CAM from weakly augmented images to obtain a reliable pseudo label, with which the prediction of strongly augmented image could be supervised. Self-training and co-training-based methods also gained popularity in data-centralized data schemes. 
However, these methods require labeled images and unlabeled images during the training process. While, in FSSL setting, the labeled and unlabeled images are decentralized to labeled and unlabeled clients, respectively. 
Instead of studying how to get a good model with labeled and unlabeled images, this paper presents a novel method on model aggregation with uneven model reliability from labeled and unlabeled clients. 


\subsection{Federated Semi-Supervised Learning}

\par FSSL can be broadly classified into two categories. One category assumes that every local client contains partially labeled images. For instance, Jeong~\etal~\cite{jeong2021federated} and Lin~\etal~\cite{lin2021semifed} let each client hold labeled and unlabeled data simultaneously. Besides, Jeong~\etal~\cite{jeong2021federated} and Zhang~\etal~\cite{zhang2020benchmarking} assume labeled data is available only at the server, and Kang~\etal~\cite{kang2020fedmvt} assume labeled and unlabeled data are isolated but inter-client sample overlapping exists. 


Another category considers that some clients are fully labeled while some clients contain unlabeled images. For example,  Liu~\etal~\cite{liu2021federated} propose to learn inter-class relationship, which is learned from labeled clients and shared among labeled and unlabeled clients. However, this method fails under the Non-IID setting, as inter-class correlations are no longer similar among clients due to data heterogeneity.
Besides, Yang~\etal~\cite{yang2021federated} introduce a consistency-based method, in which different augmentations were applied to unlabeled images with their predictions similarity maximized. While the consistency loss still works with heterogeneous data, only one unlabeled client was involved in their method. 
However, we found that these methods fail to generalize to the Non-IID setting. Our proposed RSCFed shows its robustness towards uneven model reliability under FSSL. 


%% file: Section3-method.tex
\section{Methodology}
\begin{figure*}[!t]
    \centering
    \includegraphics[width=0.8\linewidth]{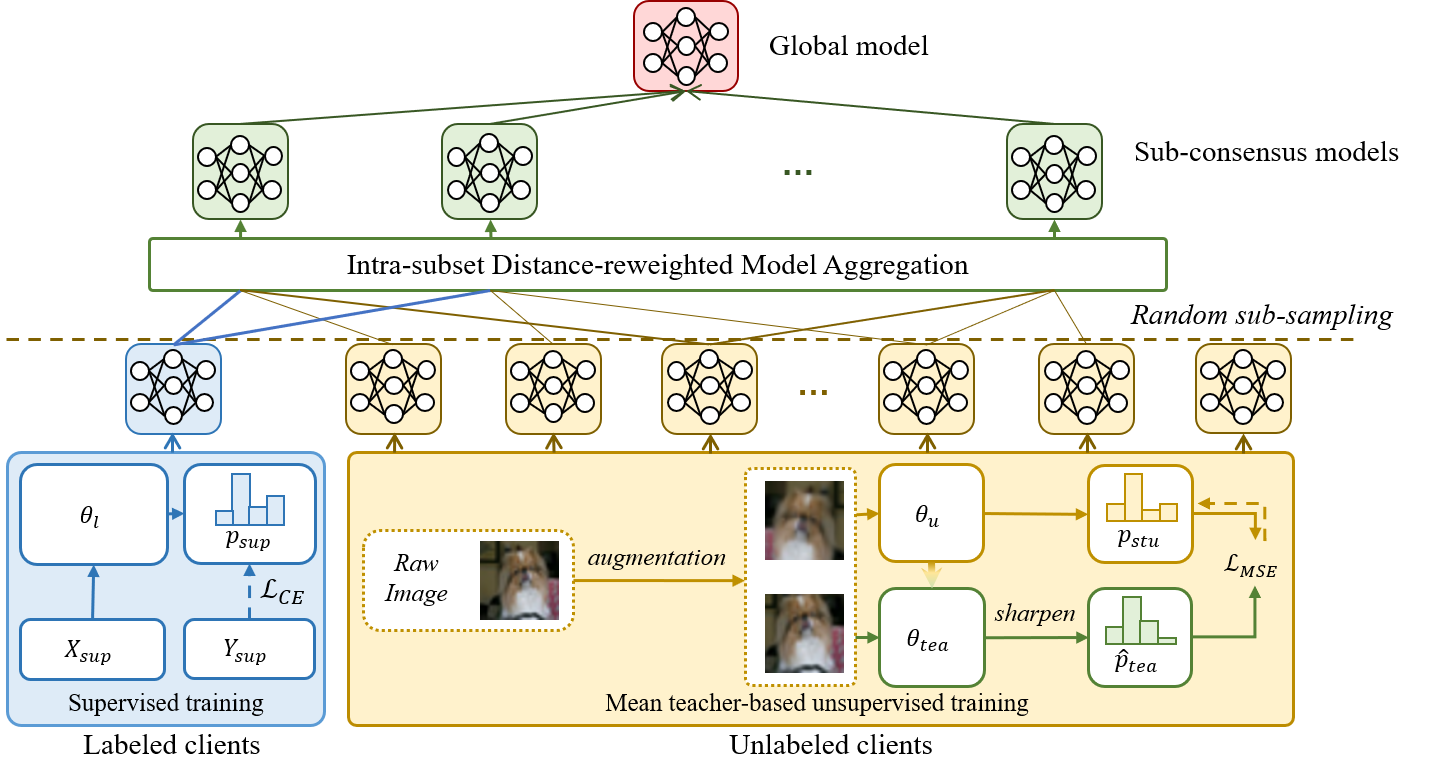}
    \vspace{-3mm} 
    \caption{An overview of our proposed RSCFed. The labeled and unlabeled clients optimize by supervised cross-entropy loss $\mathcal{L}_{CE}$ and mean-teacher-based consistency loss $\mathcal{L}_{MSE}$, respectively. Our RSCFed performs multiple random sub-sampling among all clients with distance-reweighted model aggregation (DMA) to increase the weights for clients that are close to the sub-consensus model and visa versa. This module can help avoid the influence of a deviated local model to the global model. }
    \label{pipeline}
    \vspace{-5mm}
\end{figure*}

\par Fig.~\ref{pipeline} shows an overview of our RSCFed. With some labeled and unlabeled local clients, our RSCFed respectively performs the following steps in each round: (1) Randomly sample local clients; (2) Assign current global model to selected clients as initialization, and conduct local training on selected clients; (3) Collect models from selected clients, execute distance-reweighted model aggregation (DMA) to obtain a sub-consensus model; (4) Repeat step (1)-(3) multiple times to obtain a set of sub-consensus models; (5) Aggregate a new model from the sub-consensus models set to be the next global model. 



\subsection{FSSL Setting} 
\xmli{In the methodology, we consider the FSSL with fully-labeled and fully-unlabeled clients.} Assume there are $m$ labeled clients denoted as $\left \{C_1,...,C_{m} \right\}$, and each of them has a local dataset, $\mathcal{D}^l$, defined as ${\mathcal{D}^l} = \left\{ {\left( {X_i^l,y_i^l} \right)} \right\}_{i = 1}^{N^l}$. Similarly, there are $n$ unlabeled clients denoted as $\left \{ C_{m+1},...,C_{m+n} \right \}$, and each has a dataset $\mathcal{D}^u$ containing $N^u$ unlabeled data ${\mathcal{D}^u} = \left\{ {\left( {X_i^u} \right)} \right\}_{i = 1}^{N^u}$. Our goal is to derive a good global model $\theta _{glob}$ by utilizing both labeled and unlabeled data in a decentralized scheme. 
\subsection{Local Training}
All local models are initialized with the current global model $\theta_{glob}^t$ at the beginning of $t^{th}$ synchronization round. Our proposed RSCFed adopts standard supervised and unsupervised training on labeled and unlabeled clients, respectively. For simplification, we default all representations in this section occur in $t^{th}$ synchronization round.

\vspace{3pt}
\noindent \textbf{Labeled clients} For local training on labeled clients, we adopt cross-entropy loss $\mathcal{L}_{CE}$ as the main objective:
\vskip -5 pt
\begin{equation} 
\mathcal{L}_{CE} =  - y_i\log (\hat{y}_i), 
\end{equation}
where $\hat{y}_i$ is the prediction of local data from the local model. The client then returns $\theta _l$ to server after training.

\vspace{3pt}
\noindent \textbf{Unlabeled clients} Unlabeled clients adopt mean-teacher-based consistency regularization framework, and regard student model as the local model. The teacher model $\theta _{tea}$ is initialized with $\theta_{glob}^0$ when this client is the first time selected. In each local iteration on unlabeled clients, a batch of input images is augmented twice and separately fed into the student and the teacher models. After their predictions $p_{stu}$ and $p_{tea}$ are generated, we utilize ``sharpening" defined in \cite{berthelot2019mixmatch} to increase the temperature of teacher's predictions:
%
\begin{equation}
    {{\hat p}_i} = \text{Sharpen}{\left( {{p_{tea}},\tau } \right)_i} = {{p_i^{\frac{1}{\tau }}} \mathord{\left/
 {\vphantom {{p_i^{\frac{1}{\tau }}} {\sum\nolimits_j {p_j^{\frac{1}{\tau }}} }}} \right.
 \kern-\nulldelimiterspace} {\sum\nolimits_j {p_j^{\frac{1}{\tau }}} }},
    \label{sharpen}
\end{equation}
where $p_i$ and ${\hat p}_i$ refer to each element in $p_{tea}$ before and after sharpening respectively, and $\tau$ is the temperature parameter. Thus $p_{tea}$ is ``sharpened" to ${\hat p}_{tea}$, and the sample is pushed away from the decision boundary to generate better targets for consistency alignment. With the two predictions of differently augmented input, the mean-square-error loss is adopted as the local objective on unlabeled clients:
\vskip -5pt
\begin{equation}
    {\mathcal{L}_{MSE}} = \left\| {{{\hat p}_{tea}} - {p_{stu}}} \right\|_2^2 
    \label{MSE}.
\end{equation}
Note that only the student model is updated via Eqn. \eqref{MSE}, and the teacher model receives student model parameters via exponential moving average after each local iteration:
\vskip -5pt
\begin{equation}
    {\theta _{tea} = \alpha {\theta _{stu}} + \left( {1 - \alpha } \right){\theta _{tea}}}, 
    \label{EMA}
\end{equation}
where $\alpha$ is the momentum parameter. The unlabeled client finally return the student model as its local model $\theta _u$.

\subsection{Random Sampling Consensus FL} 
We propose RSCFed, a novel FSSL framework with random subset sampling and distance-reweight model aggregation, to obtain a more robust global model from heavily biased local models. 
To be more specific, we randomly sub-sample over all clients and collect models they uploaded to dig their underlying consensus. 
Then, we obtain a sub-consensus model by aggregating collected models, where a distance-reweighted model aggregation (DMA) strategy is introduced to dynamically adjust their weights. We repeat these two steps for $M$ times to obtain a set of sub-consensus models.
Finally, we aggregate the sub-consensus models set to obtain a global model in each round.


\vspace{2pt}
\noindent \textbf{Multiple Random Sub-sampling.}
Random sub-sampling is proposed to distill a sub-consensus model. We propose to perform multiple random sub-sampling to get multiple sub-consensus models. To achieve it, at the beginning of the synchronization round $t$, we perform $M$ times independent random subsampling to sample $K$ clients.
The server then sends global model $\theta_{glob}^t$ to sampled clients, followed by executing local training on sampled clients. 
Note that if the clients are sampled multiple times in a round, we do not need to send the global model for initialization again to save communication costs.

\vspace{2pt}
\noindent \textbf{Distance-reweighted Model Aggregation.}
To enhance the robustness of the sub-consensus model, 
instead of aggregating multiple selected clients like FedAvg~\cite{mcmahan2017communication}, we propose a novel distance-reweighted model aggregation (DMA). 
Our key idea is to dynamically increase the weights for models that are close to the average model and vice versa. 
For local models of sampled clients, we perform model aggregation with a model distance-based re-weighting strategy we design. For each subset, we firstly compute an intra-subset averaged model $\theta_{avg}$:
\vspace{-5pt}
\begin{gather}
    {N_{total}} = \sum\limits_{i = 1}^K {{N_i}}, \; \text{and} \; {\theta _{avg}} = \sum\limits_{i = 1}^K {\frac{{{N_i}}}{{{N_{total}}}}  {\theta _i}}\label{tempavg},
\end{gather}

where $\theta_i$ represents the $i^{th}$ local model of the subset,  $N_i$ stands for its local data amount, and $K$ denotes the number of clients in a subset. 
Instead of simply averaging local clients, our DMA dynamically scales $w_i$ for $i^{th}$ client in each subset, as follows: 
\vskip -12pt
\begin{equation}
\small
\vspace{1 pt}
    {w_i} = \dfrac{{{N_i}}}{{{N_{total}}}}\exp \left( { - \beta  \cdot \dfrac{{{{\left\| {{\theta _i} - {\theta _{avg}}} \right\|}_2}}}{{{N_i}}}} \right), \; \text{and} \; 
    {\bar{w}_i} = \dfrac{{{w_i}}}{{\sum_j {{w_j}} }},
\label{reweight}
\end{equation}
where $\beta$ is a hyper-parameter and ${{{\left\| {{\theta _i} - {\theta _{avg}}} \right\|}_2}}$ refers to $L_2$ Norm of the model gradient between $i^{th}$ local model and temporal averaged model within the subset. The model distance is divided by local data quantity $n_i$ to reduce the impact of local iterations on model drift. We then normalize the intra-subset model weight to [0, 1]. 
\par After obtaining a set of sub-consensus models, we denote their equally weighted average to be the final global model $\theta_{glob}$: 
\vskip -10pt
\begin{equation}
  {\theta _{glob}^{t+1}} = \frac{1}{M}\sum\limits_{m = 0}^{M - 1} {\theta _{sub}^m},
\end{equation}
where $\theta _{sub}^m$ denotes the $m^{th}$ sub-consensus model. Then ${t+1}^{th}$ synchronization round is executed with $\theta _{glob}^{t+1}$ as initialization. The whole updating in $t^{th}$ synchronization round of our RSCFed is presented in Algorithm \ref{alg1}.
\begin{algorithm}[!t]
    \caption{The RSCFed framework}\label{alg1}
    \KwIn{
    $\theta_{glob}^{t}$: the global model from $t-1^{th}$ round;
    \qquad$N$: number of clients; $M$: number of subsets; $K$: number of clients in each subset}
    \KwOut{$\theta_{glob}^{t+1}$ from $t^{th}$ round}
        \For{$m\leftarrow 0 $ \KwTo $M$}{
            Randomly select $\left\{ {{C_i}} \right\}_{i = 1}^K$ from $N$ clients\newline
            \For{$k\leftarrow 0 $ \KwTo $K$}{
                send global model $\theta_{glob}$ to $C_k$\;
                $\theta_k\leftarrow\textbf{LocalTraining}(k,\theta_{glob})$
            }
            $\bar \theta\leftarrow \textbf{Avg}(\theta_k, k = 0 to K-1)$ Eqn. \eqref{tempavg}\;
            ${{\bar w}_k}\leftarrow \textbf{ReWeight}(\theta_k^m, \bar \theta^m)$ Eqn. \eqref{reweight}\;
            $\theta _{sub}^m \leftarrow \sum\limits_{k = 0}^{K - 1} {{{\bar w}_k}{\theta _k^m}}$\newline
        }
        Return $\theta _{glob}^{t + 1} \leftarrow \frac{1}{M}\sum\limits_{m = 0}^{M - 1} {\theta _{sub}^m}$
\end{algorithm}
\setlength{\textfloatsep}{10pt}

%% file: Section4-Experiments.tex
\vspace{-1mm}
\section{Experiments}
\vspace{-1mm}
To demonstrate the effectiveness and robustness of our proposed RSCFed, we conduct experiments on 3 benchmark datasets, and further evaluate RSCFed under intensive settings like different unlabeled data ratio, limited communication cost, \etc.
\vspace{-1mm}
\subsection{Dataset and Experimental Setup}
\vspace{-2mm}

\begin{table*}[!t]
\centering
\setlength{\tabcolsep}{10pt} 
\caption{Results on SVHN, CIFAR-100, and ISIC 2018 datasets under heterogeneous data partition. Note that FedIRM~\cite{liu2021federated} and Fed-Consist\cite{yang2021federated} fail to generalize in Non-IID setting. The results reported in this Table are performed with weight adjusting; see Fig.~\ref{SVHN_compare}} 
\vspace{-3mm}
\scalebox{0.8}{ 
\begin{tabular}{cc|cccccc}
\toprule
\multicolumn{1}{c|}{\multirow{2}{*}{Labeling Strategy}} & \multirow{2}{*}{Method} & \multicolumn{2}{c|}{Client Num.}         & \multicolumn{4}{c}{Metrics}                                                                                                \\ \cline{3-8} 
\multicolumn{1}{c|}{}                                   &                         & labeled & \multicolumn{1}{c|}{unlabeled} & \multicolumn{1}{c|}{Acc. (\%)} & \multicolumn{1}{c|}{AUC. (\%)} & \multicolumn{1}{c|}{Precision (\%)} & Recall (\%) \\ \hline\hline
                                                        &                         & \multicolumn{6}{c}{Dataset 1: SVHN}                                                                                                                                   \\ \hline
\multicolumn{1}{c|}{\multirow{2}{*}{Fully supervised}}  & FedAvg~\cite{mcmahan2017communication} (upper-bound)& 10& \multicolumn{1}{c|}{0}  & \multicolumn{1}{c|}{82.05}     & \multicolumn{1}{c|}{97.82}      & \multicolumn{1}{c|}{81.59}            & 77.90            \\
\multicolumn{1}{c|}{}                                   & FedAvg~\cite{mcmahan2017communication} (lower-bound)& 1 & \multicolumn{1}{c|}{0} & \multicolumn{1}{c|}{60.54}     & \multicolumn{1}{c|}{91.23}      & \multicolumn{1}{c|}{64.38}            & 57.34            \\ \hline
\multicolumn{1}{c|}{\multirow{3}{*}{Semi supervised}}   &
FedIRM~\cite{liu2021federated}         & 1 & \multicolumn{1}{c|}{9} & \multicolumn{1}{c|}{55.69}     & \multicolumn{1}{c|}{91.19}      & \multicolumn{1}{c|}{66.78}            & 56.40            \\
\multicolumn{1}{c|}{}                                   & Fed-Consist\cite{yang2021federated}   & 1 & \multicolumn{1}{c|}{9} & \multicolumn{1}{c|}{66.94}     & \multicolumn{1}{c|}{94.19}      & \multicolumn{1}{c|}{68.92}            & 66.75            \\ \cline{2-8} 
\multicolumn{1}{c|}{}                                   & \textbf{RSCFed (ours)}           & 1       & \multicolumn{1}{c|}{9}         & \multicolumn{1}{c|}{\textbf{70.26}}& \multicolumn{1}{c|}{\textbf{95.54}}& \multicolumn{1}{c|}{\textbf{73.36}}   & \textbf{68.46}   \\ \hline\hline
                                                        &                         & \multicolumn{6}{c}{Dataset 2: CIFAR-100}                                                                                                                              \\ \hline
\multicolumn{1}{c|}{\multirow{2}{*}{Fully supervised }}           & FedAvg~\cite{mcmahan2017communication} (upper-bound)& 10& \multicolumn{1}{c|}{0} & \multicolumn{1}{c|}{25.87}     & \multicolumn{1}{c|}{90.44}      & \multicolumn{1}{c|}{29.97}            & 26.01                 \\
\multicolumn{1}{c|}{}                                   & FedAvg~\cite{mcmahan2017communication} (lower-bound)& 1 & \multicolumn{1}{c|}{0} & \multicolumn{1}{c|}{12.02}     & \multicolumn{1}{c|}{76.03}      & \multicolumn{1}{c|}{10.76}            & 11.58                 \\ \hline
\multicolumn{1}{c|}{\multirow{3}{*}{Semi supervised}}            & FedIRM~\cite{liu2021federated}        & 1 & \multicolumn{1}{c|}{9} & \multicolumn{1}{c|}{14.11}     & \multicolumn{1}{c|}{79.22}      & \multicolumn{1}{c|}{14.64}            & 14.03                 \\
\multicolumn{1}{c|}{}                                   & Fed-Consist\cite{yang2021federated}   & 1 & \multicolumn{1}{c|}{9} & \multicolumn{1}{c|}{13.89}     & \multicolumn{1}{c|}{78.31}      & \multicolumn{1}{c|}{15.12}            & 12.95                 \\ \cline{2-8} 
\multicolumn{1}{c|}{}                                   & \textbf{RSCFed (ours)}                         & 1 & \multicolumn{1}{c|}{9} & \multicolumn{1}{c|}{\textbf{15.82}}& \multicolumn{1}{c|}{\textbf{81.41}}      & \multicolumn{1}{c|}{\textbf{15.85}}   & \textbf{16.37}                 \\ \hline\hline
                                                        &                         & \multicolumn{6}{c}{Dataset 3: ISIC 2018: Skin Lesion Classification}                                                                                                  \\ \hline
\multicolumn{1}{c|}{\multirow{2}{*}{Fully supervised}}  & FedAvg~\cite{mcmahan2017communication} (upper-bound)& 10& \multicolumn{1}{c|}{0} & \multicolumn{1}{c|}{84.07}          & \multicolumn{1}{c|}{95.64}          & \multicolumn{1}{c|}{76.68}                 & 62.97                 \\
\multicolumn{1}{c|}{}                                   & FedAvg~\cite{mcmahan2017communication} (lower-bound)& 1 & \multicolumn{1}{c|}{0} & \multicolumn{1}{c|}{68.14}     & \multicolumn{1}{c|}{84.12}      & \multicolumn{1}{c|}{41.91}                 & 38.61                 \\ \hline
\multicolumn{1}{c|}{\multirow{3}{*}{Semi supervised}}            & FedIRM~\cite{liu2021federated}       & 1 & \multicolumn{1}{c|}{9} & \multicolumn{1}{c|}{68.10}     & \multicolumn{1}{c|}{84.11}      & \multicolumn{1}{c|}{41.96}                 & 38.94                 \\
\multicolumn{1}{c|}{}                                   & Fed-Consist\cite{yang2021federated}   & 1 & \multicolumn{1}{c|}{9} & \multicolumn{1}{c|}{68.74}     & \multicolumn{1}{c|}{84.71}      & \multicolumn{1}{c|}{41.91}                 & 38.63                 \\ \cline{2-8} 
\multicolumn{1}{c|}{}                                   & \textbf{RSCFed (ours)}                         & 1 & \multicolumn{1}{c|}{9} & \multicolumn{1}{c|}{\textbf{70.26}}     & \multicolumn{1}{c|}{\textbf{86.01}}      & \multicolumn{1}{c|}{\textbf{45.65}}                 & 37.91                 \\ \bottomrule
\end{tabular}}
\label{leaderboard}
\vspace{-2mm}
\end{table*}
\vspace{1mm}
\noindent
\textbf{Benchmark Datasets.} We evaluate the effectiveness of our proposed method on two natural image classification datasets, i.e., SVHN and CIFAR-100. 
Moreover, to simulate the realistic privacy data decentralized-distributed scenario, we evaluate our method on ISIC 2018 (Skin Lesion Analysis Towards Melanoma Detection) consisting of 10,015 dermoscopy images with seven types of skin lesions. 
For all three benchmark datasets, 80\% images of each dataset are randomly selected for training, and the remaining images are for testing.
For SVNH and CIFAR-100, we resize the original 32$\times$32 images of these two datasets to 40$\times$40 pixels, randomly crop a 32$\times$32 region, and then utilize a normalization operation on the cropped region to generate the input of our network. 
Regarding ISIC 2018, we resize the spatial resolution of the original image from 600$\times$450 to 240$\times$240, randomly crop a 224$\times$224 region, and normalize the cropped region as the network input.

\vspace{2pt}
\noindent
\textbf{Feature extraction backbone.} When training on SVHN and CIFAR-100, we follow~\cite{li2021model} to employ a simple CNN as the feature extraction backbone, which contains two 5$\times$5 convolution layers, a 2$\times$2 max-pooling layer, and two fully-connected layers.
For the ISIC 2018 dataset, we utilize ResNet-18\cite{he2016deep} as the feature extraction backbone. 
After that, we employ a two-layer MLP and a fully-connected layer to formulate a classification network at each client for all datasets. 
Moreover, the same classification network is also utilized at each client of the compared methods for a fair comparison. 



\vspace{2pt}
\xmli{
\noindent \textbf{Federated Learning setting.} 
We follow existing methods~\cite{li2021model,Wang2020Federated,yurochkin2019bayesian} to use a Dirichlet distribution $Dir(\gamma)$ ($\gamma$=0.8 for all three benchmark datasets) to generate the non-IID data partition in clients.
After such a Non-IID data partition strategy, the number of classes and samples at each client differ from each other, and thus not all clients contain samples from all classes.  
}

\vspace{2pt}
\noindent \textbf{Implementation Details.} We utilize the SGD optimizer, and implement our method with PyTorch.
The learning rates in the labeled client and the unlabeled clients are empirically set to 0.03 and 0.021 for all methods on SVHN and CIFAR-100, and 0.002 and 0.001 for ISIC 2018. 
The batch size is set to 64 for SVHN and CIFAR-100, and 12 for ISIC 2018. 
We train 1000 synchronization rounds for all datasets to make the global model stably converged, and the local training epoch is set to 1. 
The number of sub-sampling operations $M$ and the number of local clients used in each sub-sampling operation $K$ are set as: $M$=3, and $K$=5.
Our method has three parameters: the momentum parameter $\alpha$ of Eqn.~\eqref{EMA}, temperature parameter $\tau$ of Eqn.~\eqref{sharpen}, and the scaling factor $\beta$ of Eqn.~\eqref{reweight}.
And we empirically set $\alpha$=0.001, $\tau$=0.5 for all three benchmark datasets. 
The scaling factor $\beta$ is set to 10,000 for SVHN and CIFAR-100, and 0.01 for ISIC 2018.



\begin{table}[!t]
\centering
\caption{Quantitative results of our method and the backbone model~\cite{yang2021federated} without the multiple sub-sampling operations and the distance-weighted aggregation mechanism on the three benchmark datasets. ``SSO'' denotes the multiple sub-sampling operation with model aggregation, while ``DMA'' represents the distance-weighted model aggregation mechanism.  }
\vspace{-3mm}
\setlength{\tabcolsep}{12pt}
\scalebox{0.75}{
\begin{tabular}{c|c|c|cc}
\toprule
  &\multirow{2}{*}{SSO} & \multirow{2}{*}{DMA} & \multicolumn{2}{c}{Metrics}              \\ 
\cline{4-5} & &  & Acc. (\%) & AUC (\%)          \\ \hline\hline
 & & & \multicolumn{2}{c}{Dataset 1: SVHN}      \\ \hline
 Basic &$\times$ &\ $\times$ & 66.94                & 94.19             \\ \hline
Basic + SSO &\checkmark & $\times$ & 69.15                & 95.2              \\ \hline
RSCFed (ours) &\checkmark & \checkmark        & \textbf{70.26}       & \textbf{95.54}             \\ \hline\hline
& &      & \multicolumn{2}{c}{Dataset 2: CIFAR-100} \\ \hline
Basic &$\times$ & $\times$ & 13.89                & 78.3              \\ \hline
Basic + SSO &\checkmark & $\times$ & 14.92                & 81.8              \\ \hline
RSCFed (ours) &\checkmark & \checkmark       & \textbf{15.82}       & \textbf{81.4}              \\ \hline\hline
 & &         & \multicolumn{2}{c}{Dataset 3: ISIC 2018} \\ \hline
Basic &$\times$ & $\times$ & 68.74                & 84.7              \\ \hline
Basic + SSO &\checkmark & $\times$ & 69.85                & 85.5              \\ \hline
RSCFed (ours) &\checkmark & \checkmark      & \textbf{70.26}       & \textbf{86.0}              \\ \bottomrule
\end{tabular}}
\vspace{1mm}
\label{ablation}
\end{table}

\vspace{-1mm}
\subsection{\xmli{Results with labeled and unlabeled clients}}

\xmli{
\noindent \textbf{FSSL setting.} 
In this setting, the training dataset contains ten clients: one labeled client with labeled images and nine unlabeled clients with only unlabeled samples. 
Furthermore, the same FSSL training dataset is utilized to train our network and state-of-the-art methods for a fair comparison.}

\noindent
\textbf{Implementation details.} Note that the original work in~\cite{yang2021federated,liu2021federated} reach very limited result with enough labeled data when all local models are aggregated via FedAvg~\cite{mcmahan2017communication}, see Fig.~\ref{SVHN_compare}.
Hence, we re-implement~\cite{yang2021federated}, try increased aggregation weight for labeled client from the set $\{20\%, 30\%, 50\%, 70\%\}$. Our experiments show that 50\% achieves the best classification accuracy.
Hence, we empirically enlarge the weight of labeled client to about 50\%, and other nine unlabeled clients share the remaining 50\% weight in each FSSL synchronization round. 
Such aggregating weight is also applied to guarantee the deep models performance when we re-implement FedIRM \cite{liu2021federated} and our RSCFed.

\noindent 
\textbf{Compared methods.} We compare our network against state-of-the-art FSSL methods, including (1) FedIRM~\cite{liu2021federated}, which computes an inter-class relationship labeled clients and utilizes it as extra supervisions for unlabeled clients; (2) Fed-Consist~\cite{yang2021federated}, which computes a consistency loss on predictions from multiple augmented inputs for unlabeled data in a mean teacher framework~\cite{tarvainen2017mean}.  
We also compare our network against FedAvg~\cite{mcmahan2017communication} trained with all $10$ labeled clients as the upper-bound classification result, and FedAvg~\cite{mcmahan2017communication} trained with all $1$ labeled clients as the lower-bound classification result; see Table~\ref{leaderboard}. 
Moreover, we introduce four widely-used metrics to compare different methods, and they are Accuracy, Area under the ROC Curve (AUC), Precision, and Recall.

\begin{table}[!t]
\footnotesize
\setlength{\tabcolsep}{2pt}
\centering
\caption{Comparison of our method (RSCFed) against FedIRM~\cite{liu2021federated} and Fed-Consist~\cite{yang2021federated} with number of labeled and unlabeled clients set to 2 and 8.}
\vspace{-3mm}
\scalebox{0.95}{
\begin{tabular}{c|cccc}
\toprule
\multirow{2}{*}{Method} & \multicolumn{4}{c}{Metrics}                                                                                    \\ \cline{2-5} 
                        & \multicolumn{1}{c|}{Acc.(\%)} & \multicolumn{1}{c|}{AUC(\%)} & \multicolumn{1}{c|}{Precision(\%)} & Recall(\%) \\ \hline\hline
FedIRM(origi.)~\cite{liu2021federated}& \multicolumn{1}{c|}{59.75}    & \multicolumn{1}{c|}{87.4}    & \multicolumn{1}{c|}{67.55}         & 55.26      \\
FedIRM(weight adju.)~\cite{liu2021federated}& \multicolumn{1}{c|}{74.10}    & \multicolumn{1}{c|}{94.8}    & \multicolumn{1}{c|}{76.49}         & 70.45      \\
Fed-Consist~\cite{yang2021federated}& \multicolumn{1}{c|}{75.52}    & \multicolumn{1}{c|}{96.4}    & \multicolumn{1}{c|}{77.75}         & 70.30      \\ \hline
\textbf{RSCFed(ours)}            & \multicolumn{1}{c|}{\textbf{76.65}} & \multicolumn{1}{c|}{\textbf{96.7}} & \multicolumn{1}{c|}{\textbf{78.61}} & \textbf{73.16}   \\ \bottomrule
\end{tabular}}
\label{labelednum_2}
\vspace{1mm}
\end{table}


\begin{table*}[!t]
\centering
\setlength{\tabcolsep}{8pt}
\caption{Ablation study on our method (RSCFed) in terms of different Unlabeled Client numbers and a comparison with a SOTA FSSL method (i.e., Fed-Consist~\cite{yang2021federated}).}
\vspace{-3mm}
\scalebox{0.85}{
\begin{tabular}{c|cc|cc|cc|cc}
\toprule
\multirow{2}{*}{Total client numbers }&\multicolumn{2}{c|}{Client splitting} & \multicolumn{2}{c|}{Fed-Consist~\cite{yang2021federated}} & \multicolumn{2}{c|}{Our RSCFed} & \multicolumn{2}{c}{Improvements} \\ \cline{2-9}
&Labeled        & Unlabeled       & Acc.(\%)      & AUC.(\%)      & Acc.(\%)    & AUC.(\%)    & Acc.(\%)        & AUC.(\%)       \\ \hline
5  &1              & 4               & 67.82         & 95.3          & 69.33       & 95.8        & \textbf{1.51}   & \textbf{0.5}   \\ \hline
10  &1              & 9               & 66.94         & 94.2          & 70.26       & 95.5        & \textbf{3.32}   & \textbf{1.3}   \\ \hline
15   &1              & 14              & 69.65         & 94.3          & 73.19       & 95.6        & \textbf{3.54}   & \textbf{1.3}   \\ \hline
25   &1              & 24              & 60.28         & 89.3          & 63.79       & 90.9        & \textbf{3.51}   & \textbf{1.6}   \\ \hline
35   &1              & 34              & 56.08         & 90.6          & 59.82       & 92.8        & \textbf{3.74}   & \textbf{2.2}   \\ \hline
50   &1              & 49              & 56.20         & 88.0          & 60.18       & 91.5        & \textbf{3.98}   & \textbf{3.5}   \\ \bottomrule
\end{tabular}}
\label{noise_proof}
\vspace{-4mm}
\end{table*}

\vspace{2pt}
\noindent 
\textbf{Quantitative comparisons.}
Table~\ref{leaderboard} reports the quantitative results of our network and state-of-the-art methods on three benchmark datasets in terms of four metrics. 
Basically, we can find that the results of the two compared FSSL methods (i.e., FedIRM~\cite{liu2021federated} and Fed-Consist~\cite{yang2021federated}) and our network are between the upper-bound results and the lower-bound result obtained by FedAvg~\cite{mcmahan2017communication} for all three benchmark datasets.
From these quantitative results, we can observe that our proposed RSCFed has a superior metric performance over all competitors on the three benchmark datasets.
Our superior performance over Fed-Consist indicates a generalization ability enhancement obtained by the aggregation strategy in our network.
Moreover, our network also outperforms FedIRM in terms of four metrics on three datasets. 
The reason behind is that the consistent assumption of inter-class relationship among clients is not correct due to non-IID data distribution on all clients in our work.

\vspace{1.5pt}
\noindent 
\textbf{Evaluation on SVHN.}
Regarding two compared methods, Fed-Consist has the best Accuracy performance of 66.94\%, the best AUC performance of 94.19\%, the best Precision performance of 68.92\%, and the best Recall performance of 66.75\%. More importantly, our method has larger metric scores than Fed-Consist, and achieves an Accuracy of 70.29\% (3.32\% improvement), an AUC of 95.54\% (1.35\% improvement), a Precision of 73.36\% (4.44\% improvement), and a Recall of 68.46\% (1.71\% improvement). 

\vspace{1.5pt}
\noindent 
\textbf{Evaluation on CIFAR-100.}
Regarding CIFAR-100, FedIRM has a larger Accuracy score of 14.11\%, and a larger AUC score of 79.22\%, and a larger Recall score of 14.03\%, while Fed-Consist has a larger Precision score of 15.12\%.
Compared to these two state-of-the-art methods, our network improves the Accuracy score from 14.11\% to 15.82\%, the AUC score from 79.22\% to 81.41\%, the Precision score from  15.12\% to 15.85\%, and improves the Recall score from 14.03\% to 16.37\%.

\vspace{1.5pt}
\noindent 
\textbf{Evaluation on ISIC 2018.}
Although Fed-Consist has a larger Recall score than our method, our method also achieves the best Accuracy score of 70.26\%, and the best AUC score of 86.01\%, and the best Precision score of 45.65\% among all three compared methods.
It indicates that our federated semi-supervised learning method has a higher classification accuracy for ISIC 2018.

\subsection{\xmli{Results with partially labeled clients}}
\xmli{
\noindent \textbf{FSSL setting.}
To better elaborate the ability of RSCFed in solving uneven model reliability, we further extend RSCFed to another line of FSSL, where all local \xmli{clients} are partially labeled, \ie, only 10\% images are labeled. For this setting, we adopt same network backbone as in the previous setting. Since all clients are partially labeled, no weight scaling operation is performed.} 


\xmli{
\noindent \textbf{Results} Table~\ref{partial_setting} shows our method and our baseline, \ie, Fed-Consist~\cite{yang2021federated} on SVHN dataset. Note that since the method in \cite{liu2021federated} requires extra supervision from fully labeled clients and cannot generalize to this setting, we do not list their results here. From Table~\ref{partial_setting} we can see that our RSCFed still outperforms Fed-Consist~\cite{yang2021federated} by more than 1\% in most metrics. To be more specific, our work made 1.47\% improvement in Accuracy, 1.29\% in Precision score, 1.38\% in Recall score, and 0.42\% in AUC score.}

\begin{table}[!t]
\footnotesize
\setlength{\tabcolsep}{2pt}
\centering
\caption{\xmli{Results with partially labeled clients on SVHN dataset.}}
\vspace{-3mm}
\begin{tabular}{c|cccc}
\toprule
\multirow{2}{*}{Method} & \multicolumn{4}{c}{Metrics}                                                                                    \\ \cline{2-5} 
                        & \multicolumn{1}{c|}{Acc.(\%)} & \multicolumn{1}{c|}{AUC(\%)} & \multicolumn{1}{c|}{Precision(\%)} & Recall(\%) \\ \hline\hline
Fed-Consist~\cite{yang2021federated}& \multicolumn{1}{c|}{77.54} & \multicolumn{1}{c|}{96.63}& \multicolumn{1}{c|}{77.90}  &74.11            \\
RSCFed(ours)            & \multicolumn{1}{c|}{79.01}     & \multicolumn{1}{c|}{97.05}        & \multicolumn{1}{c|}{79.19}  &75.49            \\ \bottomrule
\end{tabular}
\label{partial_setting}
\vspace{1mm}
\end{table}

\vspace{-1mm}
\subsection{Ablation Studies}
\vspace{-1mm}
We further conduct ablative experiments to evaluate the effectiveness of the major components (sub-sampling and aggregation strategy) of our RSCFed
, and further discuss its performance in terms of different unlabeled ratio, different communication cost limitations, and different hyper-parameters. All experimental results in this section are evaluated on SVHN dataset unless separately clarified.

%
\vspace{1.5pt}
\noindent \textbf{Effectiveness of SSO and DMA.}
To evaluate the effectiveness of the multiple sub-sampling operations (SSO) and the distance-reweighted model aggregation (DMA), we perform an ablation study on three benchmark datasets.
Table~\ref{ablation} compares the Accuracy and AUC scores of  quantitative results of our method and two baseline networks (i.e., ``Basic+SSO'' and ``Basic''). 
From these quantitative results, we can find that SSO and DMA have significant contributions to the success of our method in FSSL scenario.
By observing the quantitative results of ``Basic+SSO'' and ``Basic'', we can find that our SSO increases the accuracy score of 2.21\% and the AUC score of 1.01\% on SVHN, the accuracy score of 1.03\% and the AUC score of 3.5\% on CIFAR-100, as well as the accuracy score of 1.11\% and the AUC score of 0.8\% on ISIC 2018.
Moreover, the DMA of our method also helps to improve the accuracy score of 1.11\% and the AUC score of 0.34\% on SVHN, the accuracy score of 0.9\% and the AUC score of -0.4\% on CIFAR-100, as well as the accuracy score of 0.41\% and the AUC score of 0.5\% on ISIC 2018, as shown in the results of our method and ``Basic+SSO''.

\begin{table}[!t]
\footnotesize
\centering
\setlength{\tabcolsep}{5pt}
\caption{Ablation study of our method (RSCFed) in terms of different communication costs and comparisons them against Fed-Consist. ``Com. cost'' (stands for communication cost) denotes as how many times as much as that of the state-of-the-art method (i.e., Fed-Consist~\cite{yang2021federated}). }
\vspace{-3mm}
\begin{tabular}{c|cc|cc}
\toprule
\multirow{2}{*}{Method} & \multicolumn{1}{c|}{\multirow{2}{*}{Client num.}} & \multirow{2}{*}{\begin{tabular}[c]{@{}c@{}}Com. cost \end{tabular}} & \multicolumn{2}{c}{Metrics}               \\ \cline{4-5} 
                        & \multicolumn{1}{c|}{}                                   &                                                                                                  & \multicolumn{1}{c|}{Acc. (\%)} & AUC (\%) \\ \hline
Fed-Consist~\cite{yang2021federated}                & 10                        & 1.0$\times$                      & 66.94                          & 94.2     \\ \hline\hline
\multirow{4}{*}{Our RSCFed}   & 8                         & 0.8$\times$                      & 68.23                          & 94.4     \\
                        & 9                         & 0.9$\times$                      & 69.25                          & 95.0     \\
                        & 10                        & 1.0$\times$                      & 69.54                          & 95.2     \\
                        & 15                        & 1.5$\times$                      & 70.26                          & 95.5     \\ \bottomrule
\end{tabular}
\label{cost}
\end{table}

\vspace{1.5pt}
\noindent \textbf{Labeled Client Ratio.} 
Note that FedIRM~\cite{liu2021federated} focuses on ten clients with $2$ labeled client and $8$ unlabeled clients as the setting of FSSL, where more labeled data is involved. 
To evaluate the performance of RSCFed with increased labeled client ratio, we compare our work with previous arts under fixed number of clients. Following FedIRM~\cite{liu2021federated}, we empirically divides the whole training data into 10 clients consists of 2 labeled client and 8 unlabeled clients. 
Table.~\ref{labelednum_2} lists extensive metrics of our method, FedIRM~\cite{liu2021federated} and Fed-Consist~\cite{yang2021federated}. Consistent improvements made by RSCFed can be observed in four metrics. Our improvement reaches 1.13\% in accuracy, 0.3\% in AUC, 0.86\% in Precision, and notably 2.86\% in Recall.

\begin{table}[!t]
\centering
\setlength{\tabcolsep}{15pt}
\caption{Ablation study results in terms of different hyper-parameter values. $M$ denotes the number of sub-sampling, $K$ represents the number of clients in each sub-sampling.}
\vspace{-3mm}
\scalebox{0.8}{
\begin{tabular}{c cc}
\toprule
Hyper-parameters & \multicolumn{2}{c}{Metrics}               \\ \hline
$M\times~K$               & \multicolumn{1}{c|}{Acc. (\%)} & AUC (\%) \\ \hline\hline
3$\times$5               & \multicolumn{1}{c|}{70.26}     & 95.5     \\ \hline
5$\times$3               & \multicolumn{1}{c|}{70.28}     & 95.1     \\ \hline
2$\times$7               & \multicolumn{1}{c|}{70.13}     & 95.4     \\ \hline
4$\times$4               & \multicolumn{1}{c|}{70.18}     & 95.2     \\ \bottomrule
\label{Table:hyper_para}
\end{tabular}}
\vspace{-1mm}
\end{table}

\vspace{1.5pt}
\noindent \textbf{Unlabeled Client Ratio.}
To evaluate our performance under different unlabeled client ratios, we conduct an ablation study to compare different federated semi-supervised learning methods in terms of different numbers of clients, where the number of labeled client is set to 1 in all methods.
Here, we consider the whole client number as  $5$, $10$, $15$, $25$, $35$, and $50$, and Table~\ref{noise_proof} reports the results of our method and Fed-Consist~\cite{yang2021federated}). 
As can be seen, our improvement of Accuracy and AUC scores over Fed-Consist are also enlarged when the number of unlabeled clients increases. 
The accuracy improvement is progressively increased from 1.51\% to 3.98\%, and the AUC improvement is from 0.5\% to 3.5\%, as the number of unlabeled clients grows from 4 to 49.


\vspace{2pt}
\noindent \textbf{Communication Cost Limitations.}
Note that the compared two methods passed ten local client models in each synchronization round, while our method considers $15$ local models since we utilize $3$ sub-sampling operations, and $5$ local clients is selected in each sub-sampling operation. 
Hence, the communication cost of our method is $1.5$ times that of  Fed-Consist~\cite{yang2021federated}.  
We conduct an ablation study experiment to evaluate our method under different communication cost limitations. 
Specifically, we consider another three cases with $8$ clients, $9$ clients, and $10$ clients, and thus the communication cost are $0.8$, $0.9$, and $1.0$ times of the Baseline's communication cost.
Table~\ref{cost} lists the Accuracy and AUC scores of our method with different communication costs and Fed-Consist~\cite{yang2021federated}.
It shows that our network with $0.8$ time of communication cost also outperforms the state-of-the-art method (Fed-Consist~\cite{yang2021federated}) in terms of Accuracy and AUC scores.

\vspace{2pt}
\noindent \textbf{Hyper-parameters.}
Note that our network has two major hyper-parameters, and they are the number ($M$)  of sub-sampling operations and the number ($K$) of local clients used in each sub-sampling operation.
Apparently, we empirically set $M=3$, and $K=5$.
Here, we conduct an ablation study to study different choices of $M$ and $K$, and report the Accuracy and AUC scores in Table~\ref{Table:hyper_para}.
From the results, we can find that the Accuracy and AUC scores are only slightly different under different $M$ and $K$ values.

%% file: Section5-conclusion.tex
\vspace{-1mm}
\section{Conclusion}
\label{sec:conclusion}
\vspace{-1mm}
This work presents an important, practical but overlooked federated learning problem: federated semi-supervised learning with Non-IID local~\xmli{clients}. Considering the uneven reliability among labeled and unlabeled clients, our key idea is that the consensus could be reached by performing multiple sub-sampling over clients. 
Instead of simply aggregating local models, we devise a sub-consensus model by randomly sub-sampling over clients and introduce a distance-reweighted model aggregation module to aggregate sub-sampled models in each synchronization round.
Experimental results on three benchmark datasets show that our network consistently outperforms state-of-the-art methods, which proves the effectiveness of our method.  

\if 1 
\vspace{8pt}
\noindent \textbf{Negative Societal Impacts and Limitations.} 
Like almost all existing federated learning techniques, our work also has a risk of a privacy leakage in the model transmission between clients and server, as recent studies~\cite{yin2021see} prove that local data can be reconstructed via stealing the uploaded local models continuously from a specific client. 
Although Table.~\ref{cost} has proved that our method with a small communication cost also outperforms state-of-the-art FSSL methods, our method still suffers from a limitation of possessing a relatively high communication cost due to repeated sub-sampling.
Hence, we take the task of reducing the communication cost as one of the future directions of our work.
\fi 